\documentclass[conference]{IEEEtran}
\usepackage{times}
\usepackage{todonotes}
\usepackage[numbers]{natbib}
\usepackage{multicol}
\usepackage{graphics,graphicx,caption,float,subcaption,booktabs,xcolor,multirow,array,color,ifthen,tabu,colortbl,dblfloatfix,url,xparse,mathtools,patchcmd,algorithm,algorithmic,amssymb,xspace,nicefrac,microtype,amsmath,amsfonts,bm,ragged2e,tikz,stackengine,etoolbox,xpatch,enumerate,xstring,setspace,tabularx,makecell,changepage,cuted,titlesec,wrapfig,tcolorbox, sidecap,comment, bbding, placeins}
\usepackage{gensymb,comment}
\usepackage[pagebackref=true,breaklinks=true,colorlinks=true,bookmarks=false,citecolor=blue]{hyperref}
\hypersetup{
colorlinks=true,
linkcolor=blue,
filecolor=magenta,      
citecolor=blue
}

\usepackage{siunitx}
\newcommand{\lzhang}[1]{\textcolor{black}{#1}}

\IEEEoverridecommandlockouts                              

\title{\LARGE \bf 
$\bm{\mathcal{M}}^{4}\text{Diffuser}$: 
$\bm{\mathcal{M}}$ulti-View Diffusion Policy with $\bm{\mathcal{M}}$anipulability-Aware Control for Robust $\bm{\mathcal{M}}$obile $\bm{\mathcal{M}}$anipulation
}

\ifdefined\isanonymous
    \author{%
        Anonymous Authors
    \thanks{Affiliations withheld for double-blind review.
    }\\
    \thanks{
    }\\
    \thanks{
    }
    }
\else
    \author{
        Ju Dong$^{1,2,3}$, 
        Lei Zhang$^{1,3\dag}$,
        Liding Zhang$^{2}$, Yao Ling$^{2}$, Yu Fu$^{2}$, Kaixin Bai$^{1,3}$, \\ 
        Zoltán-Csaba Márton$^{3}$,
        Zhenshan Bing$^{2}$, Zhaopeng Chen$^{3}$, Alois Christian Knoll$^{2}$, Jianwei Zhang$^{1}$ 
        \thanks{\dag Corresponding author. lei.zhang-1@studium.uni-hamburg.de}
        \thanks{$^{1}$TAMS (Technical Aspects of Multimodal Systems), Department of Informatics, University of Hamburg, Hamburg, Germany. 
        }
        \thanks{$^{2}$Technical University of Munich, Germany. 
        }
        \thanks{$^{3}$Agile Robots SE, Munich, Germany. 
        }
        \thanks{This work is supported by National Key Research and Development Program of China (2025YFE0217000).
        }
    
    }
\fi

\begin{document}
    \def\@onedot{\ifx\@let@token.\else.\null\fi\xspace}
    \DeclareRobustCommand\onedot{\futurelet\@let@token\@onedot}
    \newcommand{\figref}[1]{Fig\onedot~\ref{#1}}
    \def\etal{\emph{et al}\onedot}
    \newcommand{\secref}[1]{Sec\onedot~\ref{#1}}
    \newcommand{\tabref}[1]{Tab\onedot~\ref{#1}}
    \newcommand\ananye[1]{\textcolor{red}{#1}}
    \makeatletter
    \let\@oldmaketitle\@maketitle
    \renewcommand{\@maketitle}{
    \vspace*{3mm}
    \@oldmaketitle
    \includegraphics[width=0.97\linewidth]{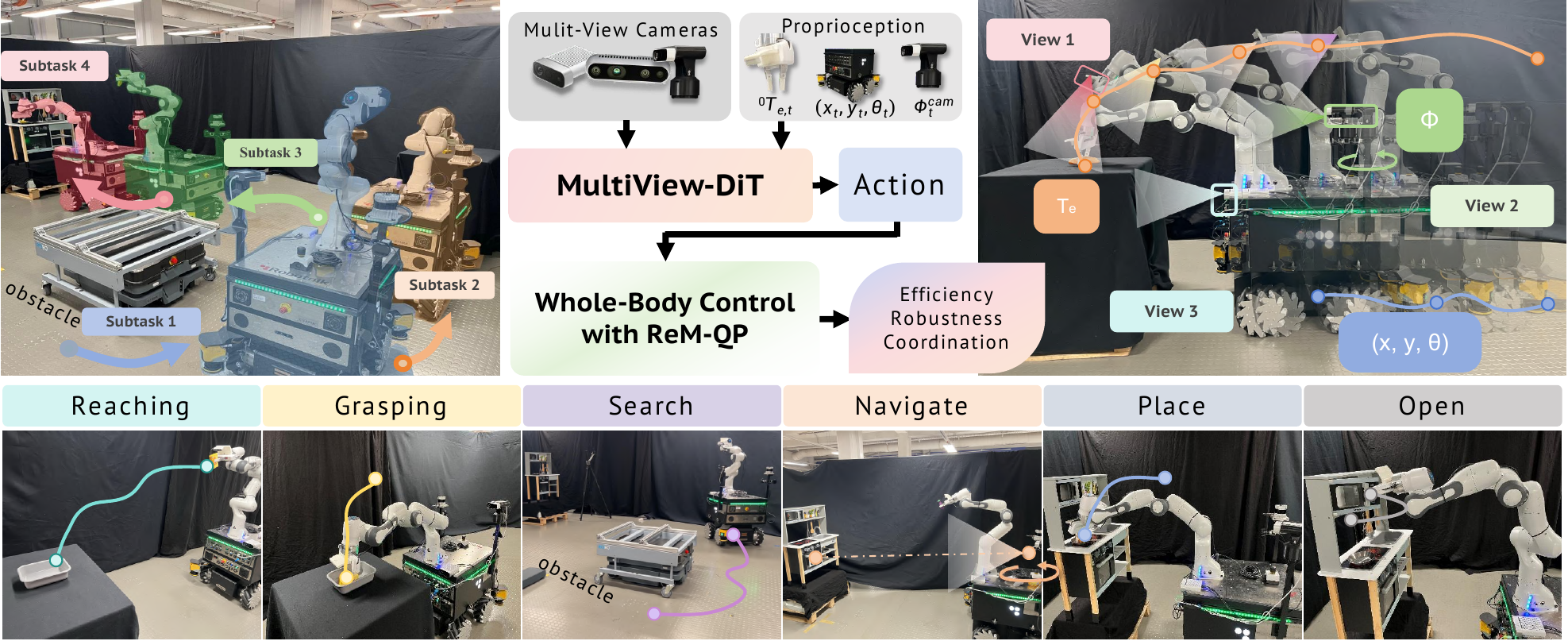}
      \centering
      \captionof{figure}{${M}^{4}\text{Diffuser}$: Multi-View Diffusion Policy and ReM-QP controller for robust whole-body mobile manipulation. 
      }
      \label{fig:teaser}
      \vspace{-0.1in}
      \bigskip}
\makeatother
\maketitle
\thispagestyle{empty}
\pagestyle{empty}


\maketitle
\thispagestyle{empty}
\pagestyle{empty}
\setcounter{figure}{1}
\begin{abstract}
Mobile manipulation requires the coordinated control of a mobile base and a robotic arm while simultaneously perceiving both global scene context and fine-grained object details. Existing single-view approaches often fail in unstructured environments due to limited fields of view, exploration, and generalization abilities. Moreover, classical controllers, although stable, struggle with efficiency and manipulability near singularities. 
To address these challenges, we propose \textbf{$\mathbf{M^{4}}\text{Diffuser}$}, a hybrid framework that integrates a Multi-View Diffusion Policy with a novel Reduced and Manipulability-aware Quadratic Programming (ReM-QP) controller for mobile manipulation. The diffusion policy leverages proprioceptive states and complementary camera perspectives with both close-range object details and global scene context to generate task-relevant end-effector goals in the world frame. These high-level goals are then executed by the ReM-QP controller, which eliminates slack variables for computational efficiency and incorporates manipulability-aware preferences for robustness near singularities. 
Comprehensive experiments in simulation and real-world environments show that \textbf{$\mathbf{M^{4}}\text{Diffuser}$} achieves \lzhang{7\%--56\%} higher success rates and reduces collisions by \lzhang{3\%--31\%} over baselines.
 Our approach demonstrates robust performance for smooth whole-body coordination, and strong generalization to unseen tasks, paving the way for reliable mobile manipulation in unstructured environments. Details of the demo and supplemental material are available on our project website~\url{https://sites.google.com/view/m4diffuser}.

\end{abstract}

\section{Introduction}
Mobile manipulation, which integrates the coordinated operation of a mobile base and a robotic arm, is a crucial capability for enabling robots to autonomously perform long-horizon tasks in real-world, unstructured environments. In recent years, notable progress has been made in this field through the development of whole-body control frameworks~\cite{sundaresan2025homer,6202432}, motion planning algorithms~\cite{wang2025ehcmmembodiedholisticcontrol}, and imitation learning strategies~\cite{sundaresan2025homer}, allowing robots to accomplish tasks such as table-to-table transfer, pick-and-place, and manipulation in kitchen scenarios. Nevertheless, several key limitations remain.

On the one hand, classical optimization-based controllers (e.g., whole-body quadratic programming controllers) provide stability and real-time performance, but they often rely on the introduction of slack variables to ensure task feasibility~\cite{ZHANG2025100207}. This design substantially increases computational overhead, reduces trajectory smoothness, and severely degrades manipulability near Jacobian singularities, frequently leading to execution failures. 

On the other hand, learning-driven methods~\cite{chi2024diffusionpolicy} demonstrate stronger adaptability and generalization across tasks. However, their deployment in practice often lacks sufficient stability. For instance, policies tend to fail when the visual input is occluded or out of view, making it difficult for the robot to maintain continuous tracking and complete the task. Single-view perception is particularly limited, as it cannot simultaneously capture both the global scene and fine-grained object details, thereby constraining the exploration ability and generalization of the policy. To address these challenges, we propose \textit{$M^{4}$}Diffuser, a hybrid control framework with two key innovations:
\begin{itemize}
\item \textbf{Multi-View Diffusion Transformer Policy:} By combining complementary perspectives, the policy can jointly capture local object details and global scene context, enabling the generation of robust end-effector goals in the world frame. Experimental results show that multi-view input substantially improves robustness and enhances generalization, achieving 23\%--56\% higher success rates and reducing collisions by 14\%--31\% compared to different diffusion policy variants.  

\item  \textbf{Reduced and Manipulability-aware  Quadratic Programming Controller (ReM-QP):} At the low-control level, we remove the slack variables of traditional QP formulations to reduce computational complexity, and introduce an inverse condition number (ICN)-based manipulability preference to ensure stability and smoothness near singularities. Simulation results demonstrate that ReM-QP reduces task execution time by \lzhang{28\%} and lowers end-effector jerk by \lzhang{35\%}, achieving a favorable balance between efficiency and smoothness.

\end{itemize}
In summary, our method achieves robust mobile manipulation in unstructured environments through the tight integration of multi-view perception and whole-body control. Across different tasks in both simulation and real-world environments, \textit{$M^{4}$}Diffuser significantly outperforms both traditional planning methods and purely learning-based approaches, \lzhang{achieving an average \textbf{+28.4\% improvement in success rate} and a \textbf{69\% reduction in collisions} compared to the OMPL baseline. Furthermore, it surpasses state-of-the-art (SOTA) methods such as HoMeR~\cite{sundaresan2025homer}, with \textbf{+10.0\% higher success rates} and \textbf{5.2\% fewer collisions}, while also demonstrating strong generalization to unseen objects and novel task configurations.}

\section{Related Work}
\label{sec:related}

\subsection{Mobile Manipulation}
In recent years, the field of mobile manipulation has made significant progress in addressing several fundamental challenges. Robots are now capable of integrating navigation and manipulation.

In the field of traditional methods, research on whole-body control frameworks~\cite{escande2014hierarchical, 6202432,ZHANG2025100207} and grasp detection algorithms~\cite{jauhri2024active} has enabled a certain degree of coordination between the mobile base and the manipulator. In addition, several works have tackled issues such as collision avoidance and adaptation to dynamic environments~\cite{10400966,burgesslimerick2022architecturereactivemobilemanipulation,Marticorena2024RMMIRM}. However, in real-world unstructured scenarios, these traditional methods still suffer from limited generalization ability, making it difficult to cope with complex and highly variable conditions.

Enhancing the searching ability is also of critical importance for mobile manipulation. Visual perception can significantly improve a robot’s searching capability~\cite{bajracharya2024demonstrating}; however, in practice, it is frequently hindered by the out-of-view problem. Traditional methods provide partial solutions to this issue, yet their effectiveness is often constrained by poor generalization performance. For example, EHC-MM~\cite{wang2025ehcmmembodiedholisticcontrol} propose a monitor-position-based servoing (MPBS) function to prevent the robot from losing track of the target during its operation.

Learning-based approaches have demonstrated stronger task and scene generalization in structured and semi-structured settings~\cite{
chen2025ac,
Yan_2025}. Recent studies have also explored learning-based coordination between the base and manipulator, for instance with diffusion policies~\cite{chen2025ac} and reinforcement learning~\cite{wenshuai2025exploiting}. Nevertheless, purely learning-driven methods typically require large-scale datasets to achieve stable performance, and the stability and reliability of the learned policies remain insufficient for industrial applications.

\subsection{Controller for Mobile Manipulation}
Classical controllers for mobile manipulation, such as hierarchical task and motion planning~\cite{escande2014hierarchical} and whole-body QP controllers~\cite{Haviland_2022}, provide strong stability and real-time performance but suffer from critical drawbacks. Holistic QP formulations often introduce slack variables to relax task constraints, which increases computational overhead and reduces trajectory smoothness in long-horizon tasks. Moreover, these optimization-based controllers are sensitive to Jacobian singularities, degrading manipulability and causing failures in cluttered environments~\cite{Vahrenkamp2012Manipulability}. Reactive whole-body controllers~\cite{6202432, 10400966} improve adaptability in dynamic settings but frequently produce jerky motions and lack global task consistency. To address these limitations, we propose the Reduced and Manipulability-aware QP (ReM-QP) controller, which eliminates slack variables for efficiency and incorporates manipulability-aware preferences for robustness.

\subsection{Hybrid Control for Mobile Manipulation}
Traditional controllers ensure stability and real-time performance but lack adaptability, while learning-based approaches offer flexibility and generalization at the cost of stability and safety. To address this, recent research has proposed hybrid control architectures that combine the stability of classical planning with the flexibility of data-driven strategies~\cite{sundaresan2025homer}. These methods have achieved promising levels of success and robustness in laboratory environments and benchmark tasks. However, our experimental results indicate that even state-of-the-art controllers still face performance bottlenecks in base–arm coordination, which ultimately limits the stability and applicability of hybrid approaches. 
Moreover, current mobile manipulation systems exhibit limited exploration ability. To overcome this limitation, we propose a multi-view policy learning framework with a manipulability-aware whole-body QP controller.

\begin{figure*}[!htb]
    \centering
    \includegraphics[width=1.0\linewidth]{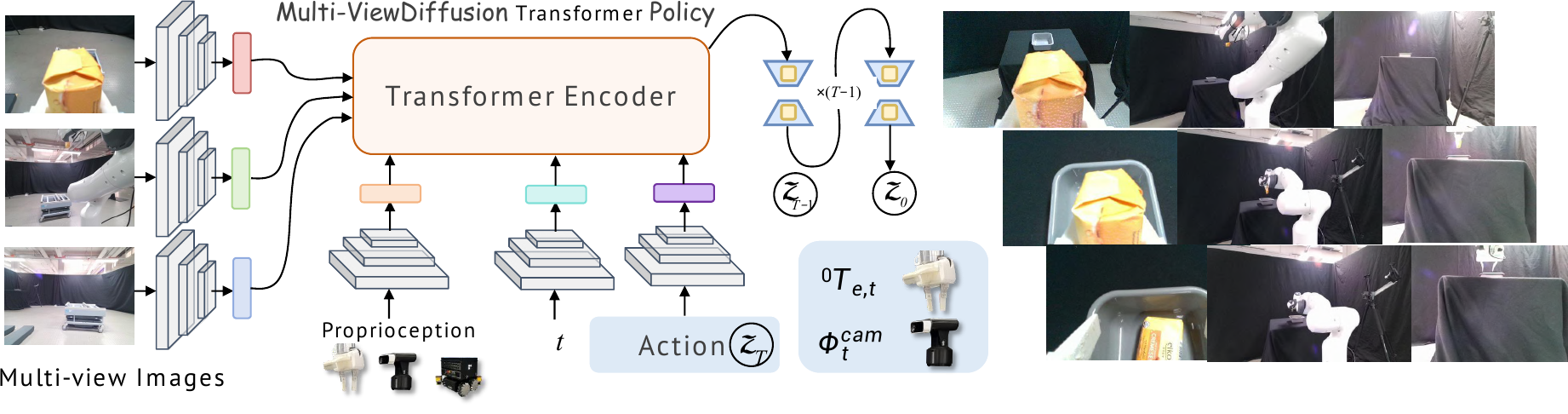}
    \caption{\small Diffusion transformer policy architecture. 
    Multi-view RGB observations and proprioceptive states 
    are encoded into latent features, which condition a denoising diffusion 
    process implemented with a Transformer. 
    The policy outputs a desired end-effector goal in the world frame, 
    which is converted into a twist and executed by the ReM-QP controller.}
    \label{fig.network_structure}
\end{figure*}

\subsection{Imitation Learning for Robotic Manipulation}
Imitation learning and generative models have been widely adopted in robotic manipulation due to its data efficiency and ease of training~\cite{zhang2025contactdexnet,wang2024tooleenet,xu2025funcanon,zhang2025responsiblerobotbench,zhang2025omnidexvlg}.  
Early approaches relied on behavior cloning and offline expert demonstrations. More recent advances, such as Diffusion Policy~\cite{chi2024diffusionpolicy} and ACT~\cite{zhao2023learningfinegrainedbimanualmanipulation}, have introduced denoising diffusion models and transformer architectures to achieve superior generalization and success rates. To collect training data for these policies, we develop a teleoperation platform for mobile manipulation~\cite{fu2024mobile}.

\section{Problem Statement and Methods}\label{sec:method}

\subsection{Problem Formulation}
We address the problem of coordinated mobile manipulation in unstructured environments. The goal is to control a mobile manipulator, comprising a holonomic mobile base and a 7-DoF robotic arm, using multi-modal observations to perform long-horizon tasks that require both global navigation and fine-grained manipulation.

At each timestep $t$, the observation is defined as:
\begin{equation}
o_t = \left\{ \text{RGB}_t^{1:3},\ \mathbf{T}_{e,t},\ \phi_t^{\text{cam}},\ (x_t, y_t, \theta_t) \right\}
\end{equation}
where:
$\text{RGB}_t^{1:3} $ denotes synchronized images from three RGB cameras. 
$\mathbf{T}_{e,t} \in SE(3)$ represents current end-effector pose. $\phi_t^{\text{cam}}$ denotes rotating rear-mounted camera angle.
$(x_t, y_t, \theta_t)$ represents mobile base pose in world coordinates. 
Given this input, the goal is to compute a joint action for the full robot body that achieves robust task execution. Our solution decomposes this into two tightly coupled modules: Multi-View Diffusion Policy and ReM-QP Controller.

The high-level Multi-View Diffusion Policy is designed to capture global scene context and local object geometry through multi-view visual inputs and proprioceptive feedback. It learns a conditional diffusion process that predicts a desired end-effector pose in the world frame:
\begin{equation}
\mathbf{T}_e^{\star} \sim \pi_{\theta}(a_t \mid o_t)
\end{equation}

The learning objective of this module is to train a conditional diffusion model that denoises perturbed actions and reconstructs the desired end-effector pose given multi-view observations and proprioceptive states. The network is introduced in Sec.~\ref{subsection:DP} and shown in Fig.~\ref{fig.network_structure}.

Whole-body controller then takes the output of the policy as input and computes the corresponding joint and base commands for execution. Formally, the process can be expressed as:
\begin{equation}
\dot{\mathbf{q}}^{\star} = \text{ReM-QP}\Big(f_{\text{servo}}\big(\mathbf{T}_{e,t}, \pi_\theta(o_t)\big)\Big),
\end{equation}

This formulation shows that the whole-body velocity command $\dot{\mathbf{q}}^{\star}$ is obtained by passing the end-effector goal $\pi_{\theta}( o_t)$ predicted by the policy through a position-based servoing law $f_{\text{servo}}(\cdot)$ and the proposed ReM-QP controller $\text{ReM-QP}(\cdot)$.

The hybrid control structure is introduced in Sec.~\ref{subsection:hybrid_control_strategy} and proposed ReM-QP controller is detailed in Sec.~\ref{subsection:ReM-QP-controller}.

 \begin{figure}[!htb]
    \centering
    \includegraphics[width=0.75\linewidth]{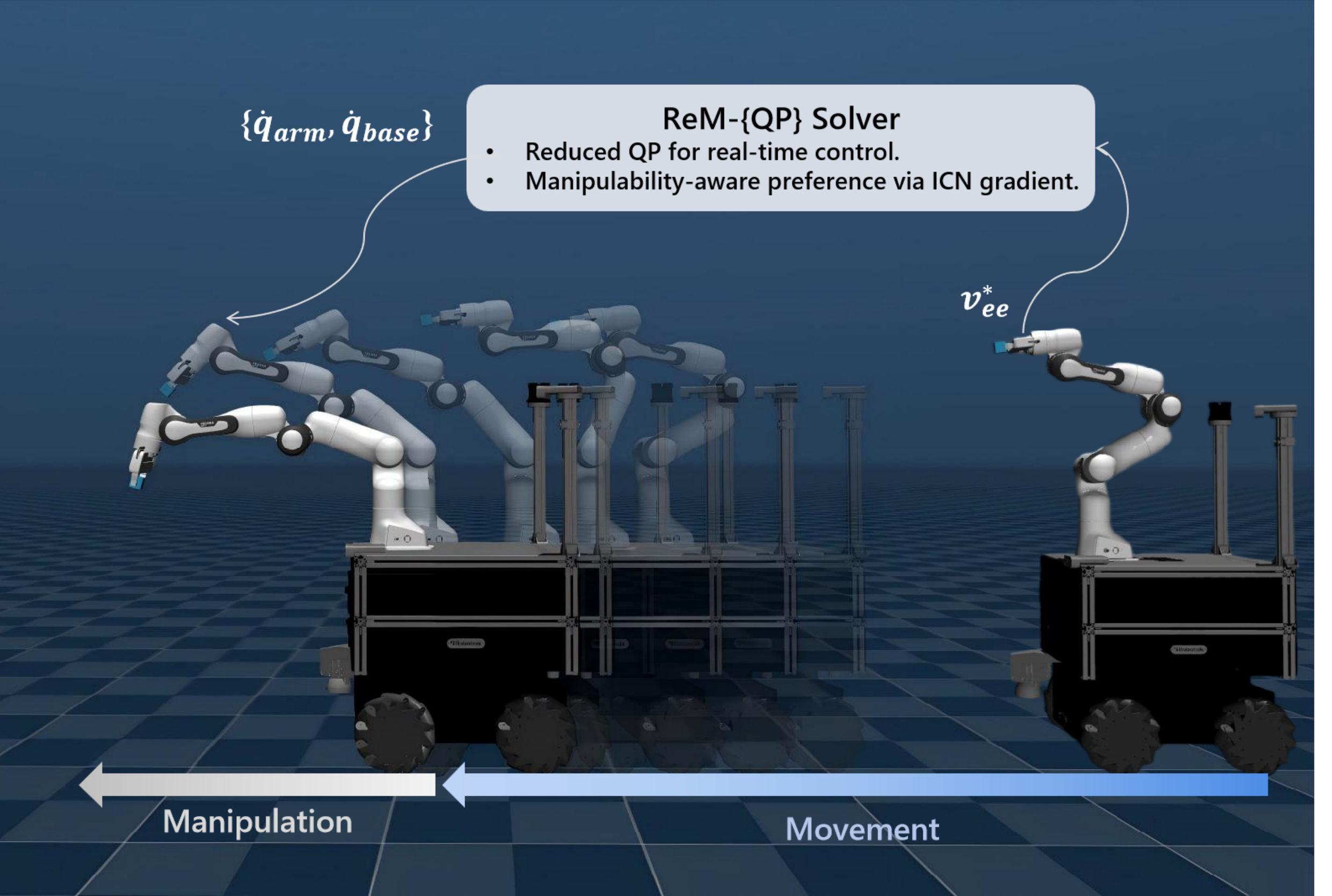}
    \caption{\small ReM-QP controller. 
    The reduced QP formulation eliminates slack variables for faster optimization, 
    while ICN-based preferences improve robustness near singularities. 
    This ensures efficient and stable whole-body execution of the high-level policy outputs.}
    \label{fig.controller}
\end{figure}

\subsection{Multi-View Diffusion Transformer Policy}
\label{subsection:DP}
Our policy network integrates a Transformer-based encoder with a conditional denoising 
diffusion process to generate robust task-space actions from raw sensory inputs\cite{dasari2024ditpi,donat2025fusingpointcloudvisual,zhao2024alohaunleashedsimplerecipe,kim2025transformerbaseddeepimitationlearning}. 
Multi-view RGB observations are first processed by convolutional backbones and concatenated 
with proprioceptive encodings. The fused features are passed through a Transformer encoder, 
which performs cross-view fusion and temporal reasoning via self-attention, producing a compact 
scene–state representation $\mathbf{h} = \text{Transformer}(\mathbf{x})$.  

Conditioned on $\mathbf{h}$, the diffusion process iteratively denoises Gaussian noise into 
feasible action distributions. Given an expert action $\mathbf{a}_0$ and diffusion time step $k$, 
the forward process is defined as
\begin{equation}
q(\mathbf{a}_k \mid \mathbf{a}_0) = \mathcal{N}\!\left(\mathbf{a}_k;\, \sqrt{\bar{\alpha}_k}\,\mathbf{a}_0,\,(1-\bar{\alpha}_k)\mathbf{I}\right),
\end{equation}
where $\bar{\alpha}_k=\prod_{s=1}^k(1-\beta_s)$ is determined by the noise schedule.  
The reverse process is modeled as
\begin{equation}
p_\theta(\mathbf{a}_{k-1} \mid \mathbf{a}_k, \mathbf{h}) 
= \mathcal{N}\!\left(\mathbf{a}_{k-1};\, 
\mu_\theta(\mathbf{a}_k, k, \mathbf{h}),\, 
\Sigma_\theta(\mathbf{a}_k, k, \mathbf{h}) \right),
\end{equation}
where $\mathbf{h}$ conditions the denoising network through cross-attention.  
The training objective reduces to noise prediction:
\begin{equation}
\mathcal{L}_{\text{diff}} = \mathbb{E}_{\mathbf{a}_0,\epsilon,k}\!\left[\|\epsilon - \epsilon_\theta(\mathbf{a}_k, k, \mathbf{h})\|^2\right].
\end{equation}

At inference, the final denoised latent is decoded into task-space commands, which are executed by a downstream whole-body controller. 

\subsection{Hybrid Control Strategy}
\label{subsection:hybrid_control_strategy}
In summary, our framework tightly couples a high-level diffusion transformer policy 
with a reduced whole-body QP (ReM-QP) controller. 
The diffusion policy leverages multi-view RGB observations and proprioceptive inputs 
to generate robust end-effector trajectories in unstructured environments. 
The ReM-QP controller then executes these high-level goals as real-time whole-body commands, 
achieving both computational efficiency (via slack elimination) 
and robustness (via ICN-based manipulability preferences), 
while ensuring safety through standard inequality constraints.  
The overall hybrid control pipeline can be expressed as:
\begin{equation}
\begin{split}
{}^0T_e^\star &\sim \pi_\theta(a_t \mid o_t), \\
\nu_{ee}^\star &= f_{\text{servo}}({}^0T_{e,t}, {}^0T_e^\star), \\
\dot q^\star &= \text{ReM-QP}(\nu_{ee}^\star).
\end{split}
\end{equation}
where the diffusion transformer policy $\pi_\theta$ outputs a desired end-effector pose ${}^0T_e^\star$, 
the servoing law converts this pose into a desired twist $\nu_{ee}^\star$, 
and the ReM-QP controller maps the twist to coordinated joint and base velocities $\dot q^\star$. 


\subsection{Reduced and Manipulability-aware Quadratic Programming (ReM-QP) Controller}
\label{subsection:ReM-QP-controller}
The ReM-QP controller serves as the low-level whole-body solver, as shown in Fig.~\ref{fig.controller}. 
Building upon the holistic QP formulation of Jesse Haviland et al.~\cite{Haviland_2022}, 
but introduces two key improvements: 
(i) elimination of slack variables for improved computational efficiency, and 
(ii) an ICN-based preference term for robust manipulability. 
These modifications yield a reduced strictly-convex QP that is both faster to solve 
and more robust near singular configurations, enabling safe execution of the high-level policy outputs.

\subsubsection{Whole-Body Kinematics}
We define four coordinate frames: the world $\mathcal{F}_0$, the mobile base $\mathcal{F}_b$, 
the arm base $\mathcal{F}_a$, and the end-effector $\mathcal{F}_e$. 
The whole-body forward kinematics is
\begin{equation}
{}^{0}T_e = {}^{0}T_b(x,y,\theta)\,{}^{b}T_a\,{}^{a}T_e(q_a),
\end{equation}
where $(x,y,\theta)$ parameterizes the holonomic base, and $q_a\in\mathbb{R}^m$ are the arm joint angles. 
The end-effector twist follows
\begin{equation}
\nu_{ee} = J(q)\,\dot q, \quad \nu_{ee} \in \mathbb{R}^6,
\end{equation}
where $\dot q \in \mathbb{R}^{n}$ concatenates base and arm velocities ($n=m+3$), 
and $J(q)\in\mathbb{R}^{6\times n}$ is the whole-body Jacobian. 
A desired twist $\nu^\star_{ee}$ is obtained from a closed-loop servoing law 
\cite{siciliano2009robotics}.

\subsubsection{Reduced QP Formulation}
The original holistic controller introduces slack variables $\delta$ to soften task constraints~\cite{wang2025ehcmmembodiedholisticcontrol,escande2014hierarchical}:
\begin{equation}
\begin{aligned}
\min_{\dot q,\,\delta} \quad & 
\frac{1}{2}
\begin{bmatrix}
\dot q \\ \delta
\end{bmatrix}^\top
\begin{bmatrix}
Q_{qq} & 0 \\
0 & Q_{\delta\delta}
\end{bmatrix}
\begin{bmatrix}
\dot q \\ \delta
\end{bmatrix}
+ c_q^\top \dot q, \\
\text{s.t.} \quad & J \dot q + \delta = \nu^\star_{ee}, \\
& A \dot q \leq b,
\end{aligned}
\label{eq:qp_slack}
\end{equation}
where $Q_{qq} \succ 0$ regularizes generalized velocities, 
$Q_{\delta\delta} \succ 0$ penalizes the slack, 
$c_q$ encodes preference terms, 
and $A \dot q \le b$ collects velocity bounds and dampers.
We eliminate the slack by substituting $\delta=\nu^\star_{ee} - J\dot q$, 
reducing the QP to a problem in $\dot q$ only:
\begin{equation}
\min_{\dot q} \; \tfrac{1}{2} \dot q^\top Q_{\text{red}} \dot q + c_{\text{red}}^\top \dot q, 
\quad \text{s.t. } A \dot q \le b,
\label{eq:qp_reduced}
\end{equation}
with
\begin{align}
Q_{\text{red}} &= Q_{qq} + J^\top Q_{\delta\delta} J, \\
c_{\text{red}} &= c_q - J^\top Q_{\delta\delta} \nu^\star_{ee}.
\end{align}
Here, $Q_{\text{red}}$ is the reduced Hessian, which combines velocity regularization
and task-tracking penalties, while $c_{\text{red}}$ is the reduced linear term. 
The latter aggregates both task-consistent preferences (e.g., manipulability) 
and the residual cost of tracking $\nu^\star_{ee}$.

\subsubsection{ICN-Based Manipulability Preference}
To enhance robustness near singular configurations, 
we introduce a preference term based on the \emph{inverse condition number (ICN)} 
of the Jacobian~\cite{Vahrenkamp2012Manipulability,10.1137/0607034}:
\begin{equation}
\text{ICN}(q) = \frac{\sigma_{\min}(J)}{\sigma_{\max}(J)} \in [0,1],
\end{equation}
where $\sigma_{\min}$ and $\sigma_{\max}$ are the smallest and largest singular values of $J$, respectively.  
The gradient of ICN is approximated via centered finite differences:
\begin{equation}
\frac{\partial \,\text{ICN}(q)}{\partial q_i} \approx 
\frac{\text{ICN}(q + \delta e_i) - \text{ICN}(q - \delta e_i)}{2\delta}, 
\quad i=1,\dots,n.
\label{eq:icn_grad}
\end{equation}
The negative gradient $-\nabla \text{ICN}(q)$ is incorporated into the linear coefficient $c_q$, 
biasing the solution toward configurations with better conditioning and away from singularities.

\subsubsection{Safety Constraints}
Based on previous method~\cite{Haviland_2022}, 
we retain standard inequality constraints $A\dot q \le b$ to encode joint velocity limits 
and velocity dampers near joint limits. 

Since the reduced Hessian $Q_{red}$ is strictly positive definite, the resulting QP remains strictly convex and admits a unique global optimum at each control cycle. The ICN-based preference only modifies the linear term $c_{red}$ and does not affect feasibility or convexity. Overall, the ReM-QP controller improves over the holistic baseline 
by eliminating slack variables for faster computation 
and introducing an ICN-based manipulability preference for robustness. 
Together with standard safety constraints, 
this yields an efficient, safe, and real-time whole-body controller 
that faithfully executes the diffusion policy outputs.

\section{Experiment}
\label{sec:experiment}
\subsection{Real-World and Simulation Experiment Setup}

\begin{figure}[htbp]
    \centering
    \includegraphics[width=0.5\linewidth]{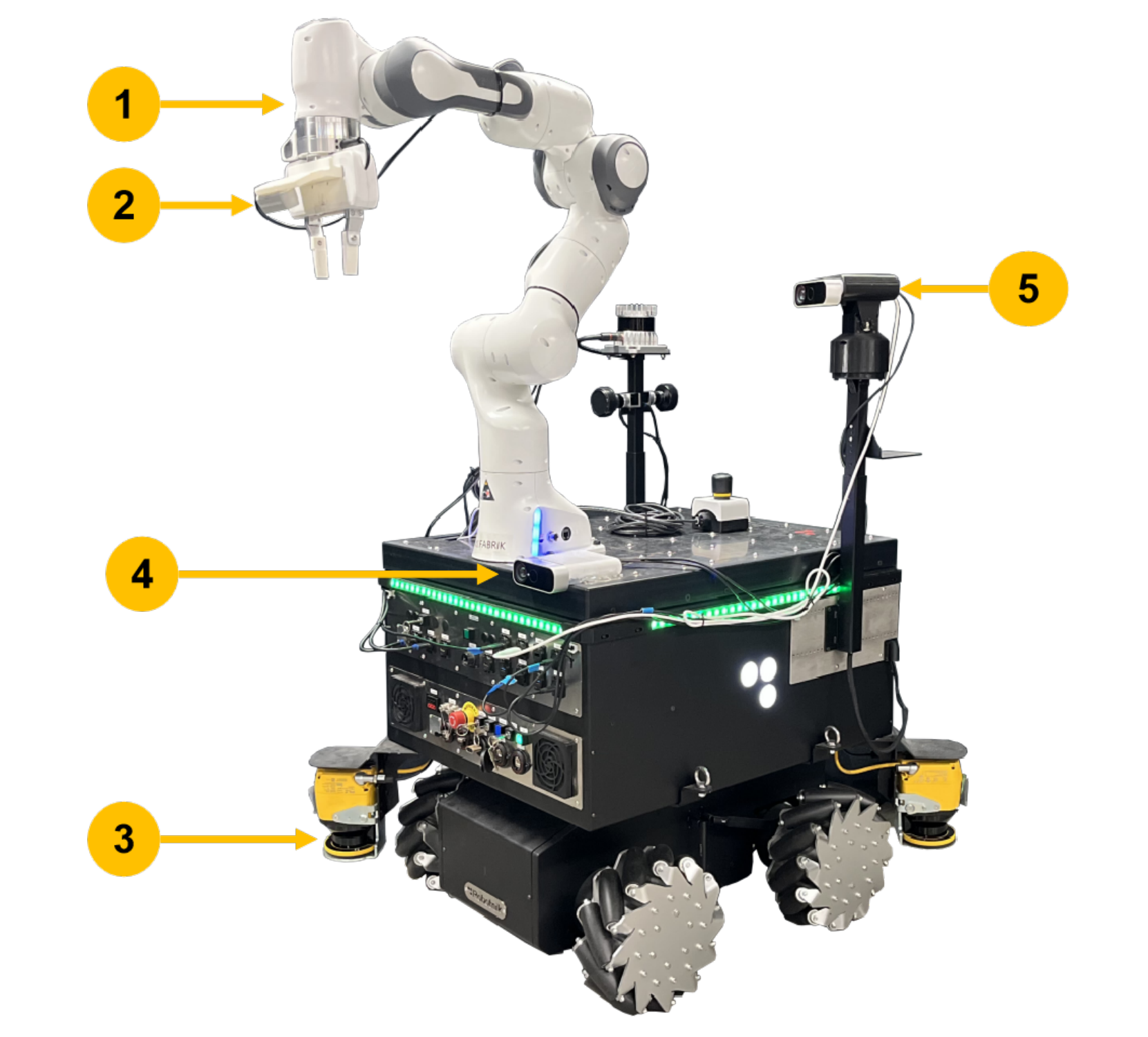}
    \caption{\small The \textit{DARKO} robot platform. It is built on an omnidirectional mobile base (RB-Kairos) with mecanum wheels, and equipped with a Franka-Emika-Panda robotic arm (1) carrying a wrist-mounted RealSense D435i camera (2). Additional sensing equipment includes two Sick MicroScan 2D lidars (3), a front-facing Azure Kinect RGB-D camera (4), and an adjustable Azure Kinect RGB-D camera (5)}
    \label{fig:DARKO}
\end{figure}

We employ the \textit{DARKO} robot for experiments, as shown in Fig.~\ref{fig:DARKO}. \lzhang{For policy training, we use teleoperation to collect 350 trajectories for each task. Implementation details of teleoperation are introduced in Supplementary Materials.} We simulate the \textit{DARKO} mobile manipulator in the MuJoCo physics engine. 
The scene consists of two tables (Table~A and Table~B) and a single graspable object initially placed on Table~A.

\subsection{Experiment Setup of Whole-Body Control in Simulation}
\begin{figure}[htbp]
    \centering
    \begin{subfigure}[t]{0.24\linewidth}
        \centering
        \includegraphics[width=\linewidth]{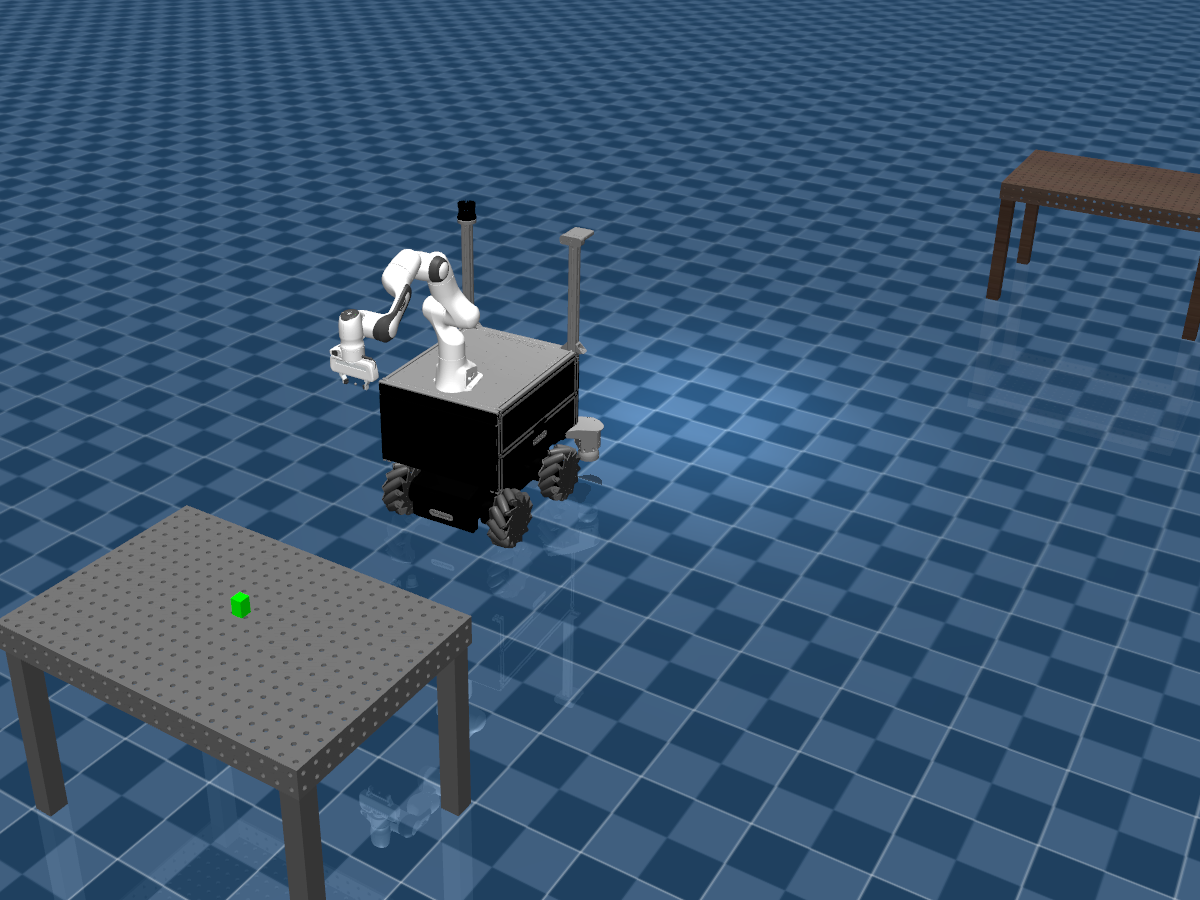}
        \caption{Nav A}
    \end{subfigure}\hfill
    \begin{subfigure}[t]{0.24\linewidth}
        \centering
        \includegraphics[width=\linewidth]{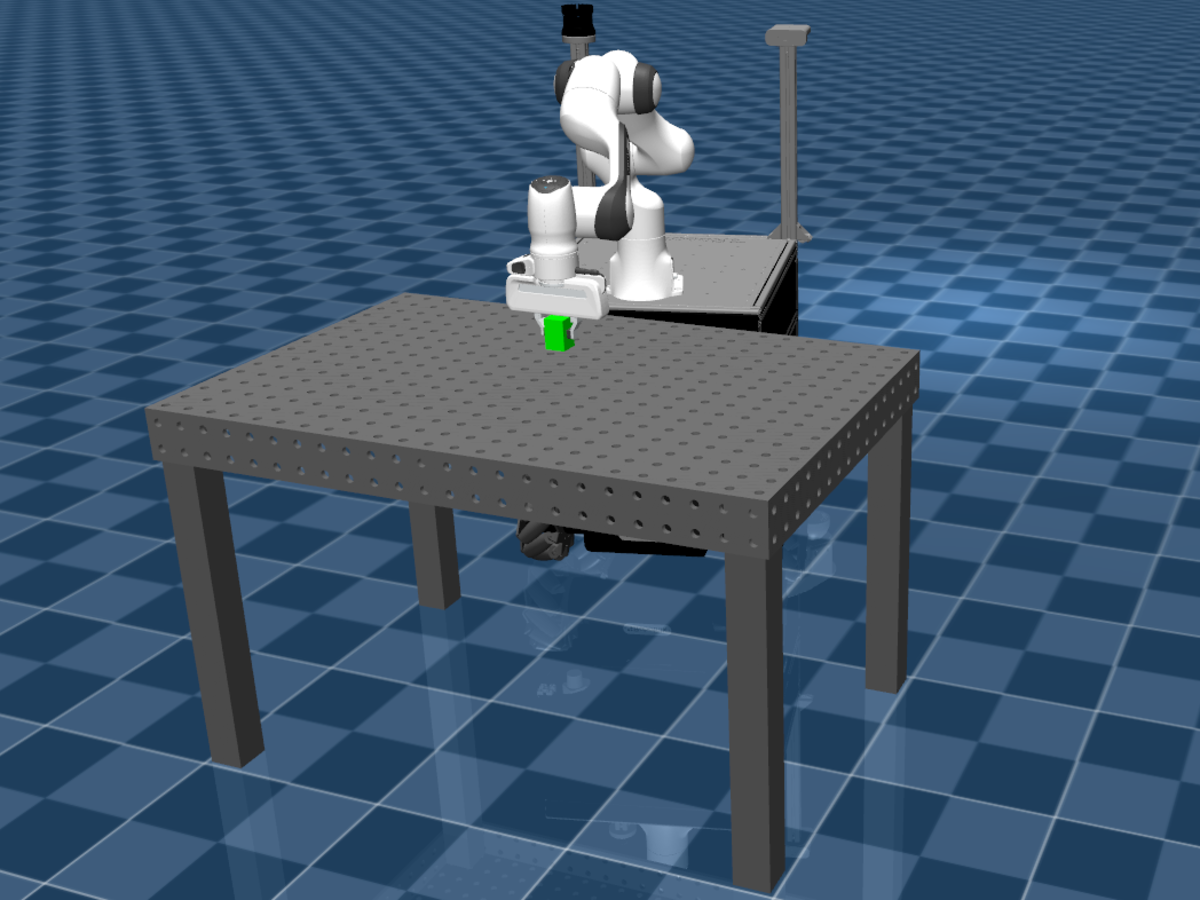}
        \caption{Desk A}
    \end{subfigure}\hfill
    \begin{subfigure}[t]{0.24\linewidth}
        \centering
        \includegraphics[width=\linewidth]{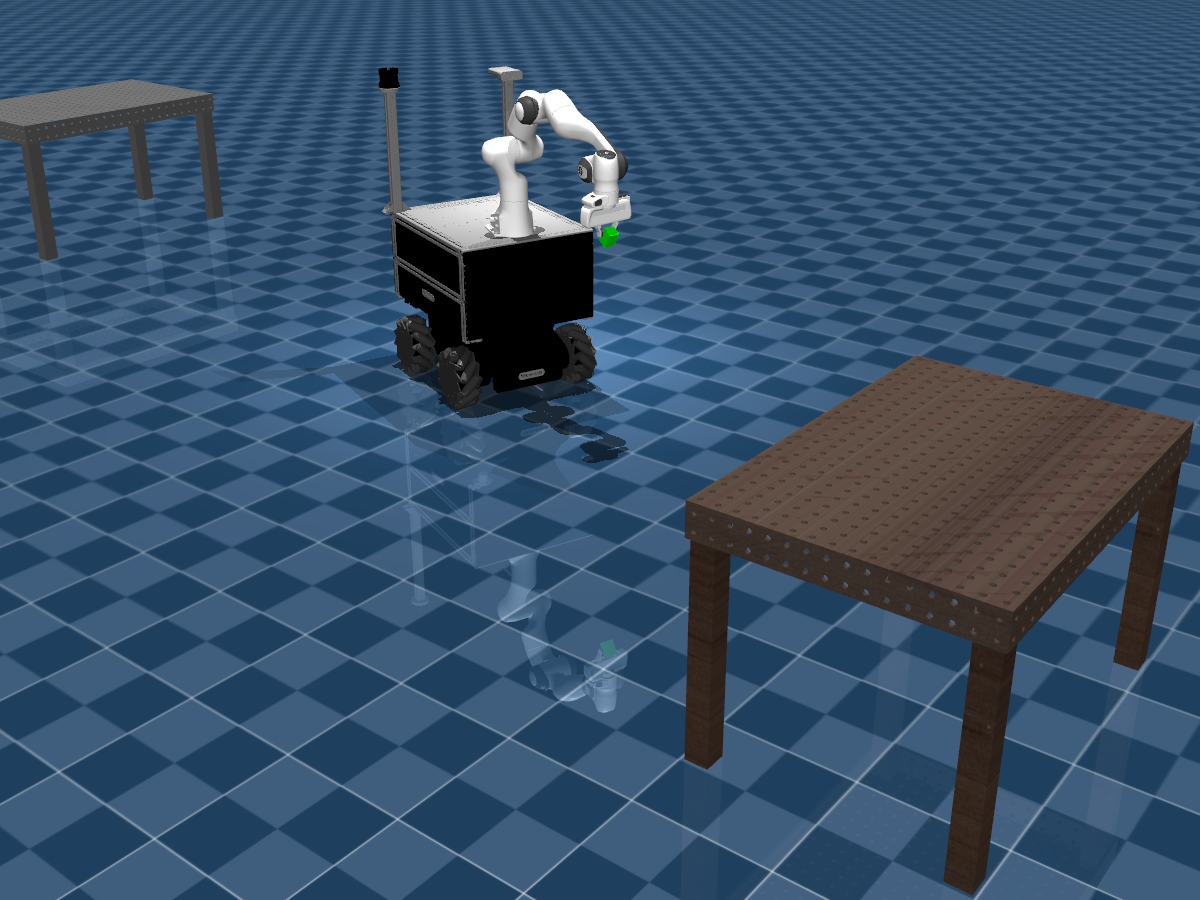}
        \caption{Nav B}
    \end{subfigure}\hfill
    \begin{subfigure}[t]{0.24\linewidth}
        \centering
        \includegraphics[width=\linewidth]{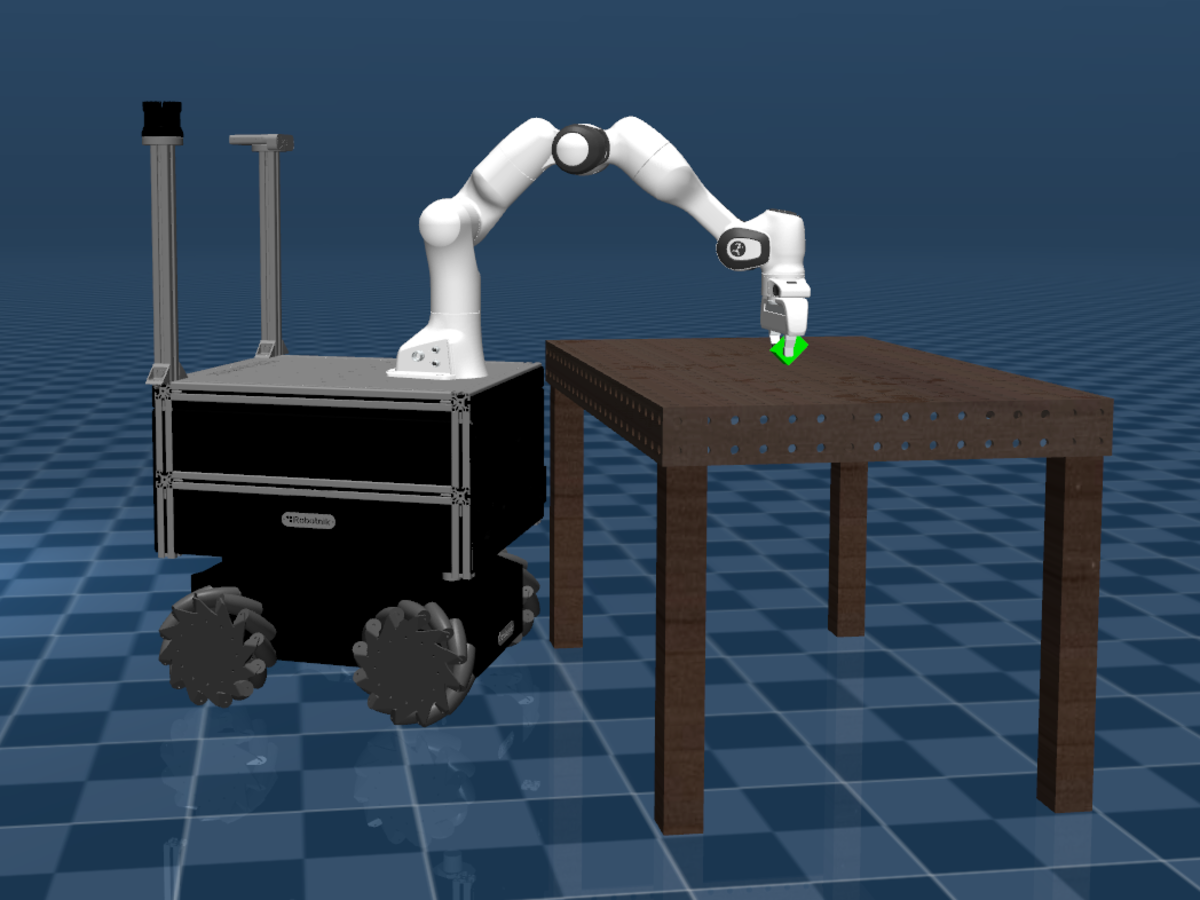}
        \caption{Desk B}
    \end{subfigure}
    \caption{\small Task phases for the mobile manipulation benchmark in the MuJoCo simulator. 
    The four phases consist of two navigation segments (Nav A, Nav B) and two dexterous manipulation segments (Desk A, Desk B).}
    \label{fig:task_phases}
\end{figure}

We evaluate proposed whole-body control (WBC) in the simulation environment. The robot is controlled to move the object from Table A to Table B. As shown in Fig.~\ref{fig:task_phases}, the task is decomposed into the following four phases: 
(i) \textbf{Nav A} — navigate to Table~A; 
(ii) \textbf{Desk A} — grasp and lift the object; 
(iii) \textbf{Nav B} — navigate to Table~B while holding the object; 
(iv) \textbf{Desk B} — place the object and lift the arm. 
This structured setup enables phase-wise comparison across controllers 
and allows us to jointly evaluate omnidirectional navigation 
and dexterous manipulation under a unified whole-body control framework. 

\textbf{Variants: } For ablation study and comparison experiment, we evaluate the following controller variants:
\begin{itemize}
    \item \textbf{Baseline}: Holistic whole-body controller with slack variables~\cite{Haviland_2022}.
    \item \textbf{Elim. Slacks}: Our reduced QP formulation that eliminates slack variables to improve computational efficiency.
    \item \textbf{Baseline+ICN}: Holistic baseline controller augmented with ICN-based grasp preference.
    \item \textbf{Ours (ReM-QP)}: Our method with slack elimination and ICN-based preferences.
\end{itemize}

\textbf{Evaluation Metrics: }We measure phase-wise execution time and global motion quality metrics, 
including RMS end-effector acceleration, RMS jerk, and total task time. 
These metrics jointly reflect both efficiency (execution time) and motion quality (trajectory smoothness).

\begin{table}[t]
\centering
\caption{\small Overall performance across controllers. 
$\downarrow$ lower is better, $\uparrow$ higher is better.}
\label{tab:overall}
\resizebox{1.0\linewidth}{!}{
\begin{tabular}{lccc}
\toprule
Method & EE Acc. RMS (m/s$^2$)$\downarrow$ & EE Jerk RMS (m/s$^3$)$\downarrow$ & Total Time (s)$\downarrow$ \\
\midrule
Baseline~\cite{Haviland_2022}          & 0.151 & 4.61 & 146.67 \\
Elim. Slacks      & 0.184 & 9.47 & \textbf{83.86} \\
Baseline+ICN      & \textbf{0.055} & \textbf{1.58} & 128.57 \\
Ours (ReM-QP)     & 0.087 & 2.98 & 105.87 \\
\bottomrule
\end{tabular}
} 
\end{table}

\begin{table}[t]
\centering
\caption{\small Per-phase execution time (s). 
Task is split into navigation (Nav A/B) and dexterous phases (Desk A/B).}
\label{tab:phases}
\resizebox{\linewidth}{!}{
\begin{tabular}{lcccc}
\toprule
Phase & Baseline & Elim. Slacks & Baseline+ICN & Ours (ReM-QP) \\
\midrule
Nav A (to Table A)   & 58.59 & 23.06 & 36.21 & 30.15  \\
Desk A (grasp+lift)  & 9.15  & 6.60 & 8.65  & 7.33 \\
Nav B (to Table B)   & 59.06 & 41.85 & 67.64 & 47.00 \\
Desk B (place+lift)  & 19.87 & 12.17 & 16.07 & 21.40 \\
\midrule
Total                & 146.67 & \textbf{83.86} & 128.57 & 105.87 \\
\bottomrule
\end{tabular}
}
\end{table}

\begin{figure}[!t]
    \centering
        \includegraphics[width=0.8\linewidth]{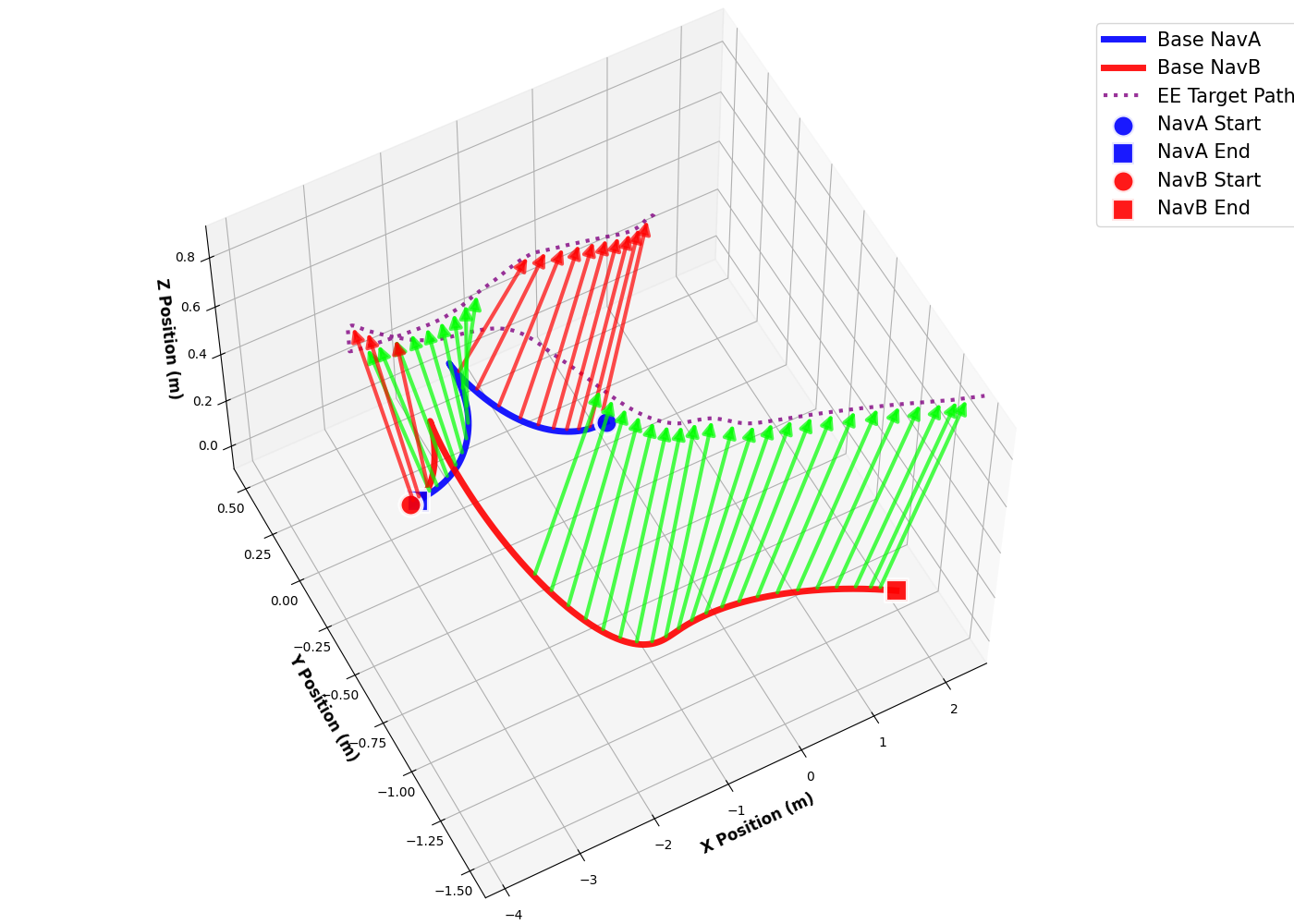}
    \caption{\small Visualization of ReM-QP whole-body control. 
    The controller ensures coordinated base--arm behavior by automatically coupling base motion to end-effector trajectories.}
    \label{fig:rem_visualization}
\end{figure}

\subsection{Qualitative Results of Whole-Body Control with ReM-QP}
To complement the quantitative metrics, we visualize the trajectories of both the end-effector 
and the mobile base. 
Fig.~\ref{fig:rem_visualization} illustrates how the ReM-QP controller couples end-effector goals 
with coordinated base movements: when the end-effector advances toward the target, 
the base follows with forward motions (green arrows), while backward adjustments (red arrows) 
emerge when the controller repositions the base to maintain reachability and manipulability. 
This visualization highlights the key property of our controller: 
\textbf{end-effector goals actively guide base motion}, enabling smooth whole-body coordination 
across both navigation and manipulation.

\subsection{Quantitative Results of Whole-Body Control with ReM-QP}
The quantitative results of whole-body control are summarized in Tab.~\ref{tab:overall} and Tab.~\ref{tab:phases}. The four controllers are compared in terms of overall motion quality (end-effector acceleration/ jerk RMS and total time) and phase-wise execution time.

As shown in Tab.~\ref{tab:overall}, the baseline controller suffers from high jerk (4.61 m/s$^3$) 
and long completion time (147 s). Eliminating slack variables (Elim. Slacks) drastically reduces 
the total execution time to 83.9 s, highlighting the efficiency gain of the reduced QP formulation. 
However, this speed-up comes at the cost of higher acceleration and jerk, 
reflecting less smooth motions. In contrast, augmenting the baseline with ICN-based preferences 
(Baseline+ICN) significantly improves smoothness with the lowest acceleration (0.055 m/s$^2$) 
and jerk (1.58 m/s$^3$). However, the overall time remains long (128.8 s) 
due to inefficiencies in navigation.

Our proposed method (ReM-QP) combines the benefits of both approaches. 
By jointly eliminating slacks and incorporating ICN-based preferences, 
it achieves a balanced performance: the total task time is reduced to 105.87 s, 
while maintaining substantially lower jerk (2.98 m/s$^3$) than Elim. Slacks. 
This indicates that ReM-QP preserves smooth and stable grasping while still being considerably faster 
than the baseline controller.

The phase-wise breakdown in Tab. \ref{tab:phases} provides further insights. 
Navigation phases (Nav A/B) dominate task time. 
Elim. Slacks yields the fastest navigation, but at the expense of motion quality. 
Baseline+ICN, on the other hand, improves dexterous phases (Desk A/B) with smoother end-effector motions, 
but shows slower navigation due to less effective base–arm coordination. 
Our method strikes a middle ground: navigation is faster than baseline and Baseline+ICN, 
while desk phases remain smooth and reliable. Notably, Desk A is completed in 7.3 s, 
close to the efficiency of Elim. Slacks, but without compromising grasp stability.

Overall, these results demonstrate that the proposed ReM-QP controller achieves a favorable trade-off: 
reducing task time by 28\% compared to the baseline while lowering jerk by 35\%, 
thus ensuring both efficiency and robustness in long-horizon mobile manipulation. 
These findings justify the use of ReM-QP as the low-level controller in the subsequent real-world \textit{$M^{4}$}Diffuser experiments.

\subsection{Experiment Setup of Real-World Mobile Manipulation}
We evaluate the \textit{$M^{4}$}Diffuser framework on the physical \textit{DARKO} robot 
across a suite of mobile manipulation tasks that require both reliable navigation 
and dexterous manipulation. 
Each task consists of two stages: (i) navigation through the scene while avoiding obstacles, 
and (ii) a tabletop manipulation task. 
We consider both simplified environments (``easy'') and realistic kitchen scenes with unstructured layouts.  
\begin{itemize}
    \item \textbf{Reaching (easy/ kitchen):} The robot grasps an object 
    and navigates around obstacles to place it into a container on a table or on the kitchen stove. 
    \item \textbf{Pick-and-Place (easy/ kitchen):} The robot navigates around obstacles, 
    grasps an object from a table, and places it into a container on the same table or delivers it to a desinated location in the kitchen setup.
    \item \textbf{Open Door (kitchen):} The robot navigates to a microwave in 
    the kitchen scene and open the door.
\end{itemize}

Specifically, LiDAR-based SLAM method is utilized to obtain the location of mobile base.

\textbf{Baselines for Comparison Experiments: }  
We compare our approach against the following baselines:  

\begin{itemize}
    \item \textbf{OMPL baseline with AprilTags:}  
    A traditional decoupled planning pipeline implemented using the OMPL library~\cite{sucan2012the-open-motion-planning-library}.  
    The mobile base is driven by LiDAR-based SLAM, while arm are planned separately in joint space.    
    \item \textbf{Diffusion policy (single-view/ multi-view):}  
    Diffusion policy is conditioned only on the single-view or multi-view information outputs joint-level actions for both the arm and the base.  

    \item \textbf{Diffusion policy (multi-view) + ReM-QP:}  
    The multi-view diffusion policy predicts high-level end-effector goals,  
    which are then converted into whole-body joint commands by our proposed ReM-QP controller.   
    \item \textbf{HoMeR~\cite{sundaresan2025homer}(multi-view) + WBC:} combines RGB-based relative actions and point cloud-based keyposes, executed via a WBC.
    \item \textbf{Diffusion Transformer (multi-view) + ReM-QP  (ours):} Multi-view diffusion transformer policy with the ReM-QP controller. 
\end{itemize}
\lzhang{For all multi-view settings except HoMeR, the rear camera is mounted on an adjustable bracket, 
allowing flexible viewpoint adaptation; HoMeR relies on fixed viewpoints.}

\subsection{Qualitative Results of Real-World Experiments}
\begin{figure*}[t]
    \centering
    \includegraphics[width=0.9\linewidth]{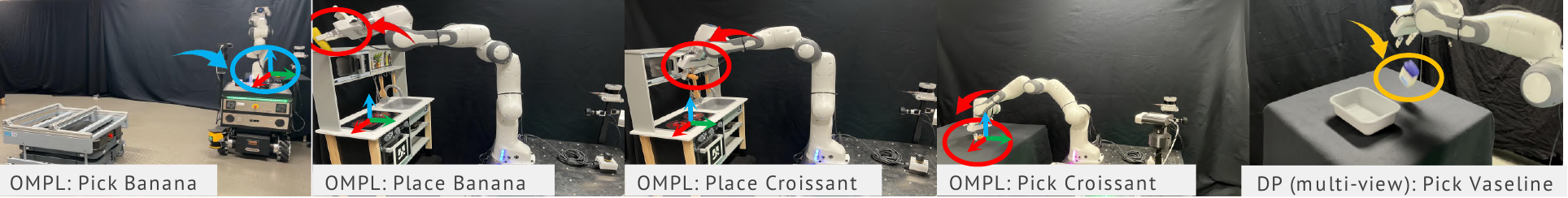}
    \caption{\small Failure cases in real-world experiments. 
    (\textbf{Case 1}) OMPL baseline fails to pick a banana: wrist camera loses AprilTag (blue arrow/circle).  
    (\textbf{Case 2}) OMPL baseline fails to place a banana: trajectory error causes collision with the cabinet edge (red arrow/circle).  
    (\textbf{Case 3}) OMPL baseline fails to place a croissant: arm collides with the cabinet door (red arrow/circle).  
    (\textbf{Case 4}) OMPL baseline fails to grasp a croissant: arm strikes the table surface (red arrow/circle).  
    (\textbf{Case 5}) DP (multi-view) fails to pick Vaseline: jitter leads to object slip (yellow arrow/box). }
    \label{fig:failure_cases}
\end{figure*}

Fig.~\ref{fig:failure_cases} illustrate representative failures in real-world rollouts. The \emph{OMPL baseline} frequently fails in kitchen scenes, mainly due to its reliance on AprilTags 
and the brittleness of decoupled planning.  
The \emph{Diffusion Policy (multi-view)} variant, although marker-free, still suffers from unstable base–arm coordination, 
leading to jitter and object slips. These qualitative examples highlight two major limitations of existing approaches: 
strong dependence on artificial markers and unstable execution without optimization. 
Both issues are addressed by our framework. These cases highlight the brittleness of marker-dependent planning and the instability of direct policy execution, 
which motivate the need for our design.

\begin{table*}[t]
\centering
\caption{\small Task-wise performance of different methods. Each task was evaluated over 50 trials.
Metrics: success rate (\%) and collision rate (\%).}
\label{tab:taskwise_extended}
\resizebox{0.9\linewidth}{!}{
\begin{tabular}{lcccccccccccc}
\toprule
\multirow{2}{*}{\textbf{Task}} 
& \multicolumn{2}{c}{OMPL baseline} 
& \multicolumn{2}{c}{DP (single-view)} 
& \multicolumn{2}{c}{DP (multi-view)} 
& \multicolumn{2}{c}{DP (multi-view) + ReM-QP} 
& \multicolumn{2}{c}{HoMeR~\cite{sundaresan2025homer}} 
& \multicolumn{2}{c}{ \textit{$M^{4}$}Diffuser (ours)} \\
\cmidrule(lr){2-3} \cmidrule(lr){4-5} \cmidrule(lr){6-7} \cmidrule(lr){8-9} \cmidrule(lr){10-11} \cmidrule(lr){12-13}
& Success & Coll. & Success & Coll. & Success & Coll. & Success & Coll. & Success & Coll. & Success & Coll. \\
\midrule
Reaching (easy)          & 66.0 & 18.0 & 32.0 & 34.0 & 68.0 & 20.0 & 82.0 & 6.0  & 82.0 & 8.0  & \textbf{90.0} & \textbf{4.0} \\
Reaching (kitchen)       & 54.0 & 22.0 & 28.0 & 38.0 & 60.0 & 22.0 & 76.0 & 8.0  & 74.0 & 14.0 & \textbf{84.0} & \textbf{6.0} \\
Pick-and-Place (easy)    & 60.0 & 20.0 & 30.0 & 36.0 & 66.0 & 18.0 & 80.0 & 10.0 & 76.0 & 10.0 & \textbf{86.0} & \textbf{6.0} \\
Pick-and-Place (kitchen) & 52.0 & 24.0 & 24.0 & 40.0 & 58.0 & 24.0 & 76.0 & 12.0 & 72.0 & 12.0 & \textbf{82.0} & \textbf{8.0} \\
Open Door (kitchen)      & 38.0 & 26.0 & 20.0 & 42.0 & 44.0 & 20.0 & 62.0 & 12.0 & 58.0 & 16.0 & \textbf{70.0} & \textbf{10.0} \\
\midrule
\textbf{Average}         & 54.0 & 22.0 & 26.8 & 38.0 & 59.2 & 20.8 & 75.2 & 9.6  & 72.4 & 12.0 & \textbf{82.4} & \textbf{6.8} \\
\bottomrule
\end{tabular}
}
\end{table*}

\subsection{Quantitative Results of Real-World Experiments}

The quantitative results are detailed in Tab.~\ref{tab:taskwise_extended}. On average, the \textbf{OMPL baseline} achieves a success rate of 54\% and a collision rate of 22\%.
In the kitchen environment, the success rate decreased while the collision rate increased, indicating the higher difficulty of tasks in this setting.
    \textbf{DP (single-view)} performs worst overall, with success rates averaging only $\sim$27\% 
    and collisions 38\%, showing that single-view perception is insufficient. \textbf{DP (multi-view)} improves robustness compared to single-view and performs on par with OMPL, 
    but collisions remain high ($\sim$21\%). For \textbf{DP (multi-view) + ReM-QP}, optimization brings clear gains: 
    success rises to $\sim$75\% and collisions drop below 10\%. \textbf{HoMeR}~\cite{sundaresan2025homer} achieves intermediate performance, 
    averaging 72.4\% success and 12\% collisions, slightly below DP (multi-view) + ReM-QP.  
    Finally, our method achieves the best overall performance, averaging \textbf{82.4\% success} 
    with only \textbf{6.8\% collisions}, and shows particularly strong improvements in challenging kitchen tasks.

These results confirm the necessity of our hybrid framework:  
OMPL is brittle and marker-dependent, DP alone is unstable, and even DP+ReM-QP is less reliable than our transformer-based hybrid.  
The ReM-QP controller is essential for translating high-level diffusion outputs into smooth, feasible whole-body motions, 
while the multi-view transformer policy provides robust spatial reasoning.   
In addition, the use of an adjustable rear-mounted camera ensures that target objects remain consistently within view, 
further improving robustness in real-world environments.

\subsection{Generalization Experiments}
To further evaluate robustness and generalization, 
we conducted additional experiments with unseen objects and altered task configurations. 
Although the policy was trained primarily on banana manipulation, 
we replaced the target with a croissant and an eggplant. 
The robot successfully grasped and placed both objects with high success rates of 80\% in 50 trials, 
demonstrating strong object-level generalization.  
Moreover, when target placement positions were shifted to novel locations, 
the hybrid method maintained reliable performance, 
showing robustness to position variations.  
These experiments highlight the adaptability of our multi-view diffusion transformer policy, 
reinforced by the ReM-QP controller, 
and underline its practical deployability in unstructured real-world environments.  
\label{sec4:expriments}

\section{Conclusion and Future Work}
\label{sec:conclusion}

This work presented a hybrid control framework \textit{$M^{4}$}Diffuser for mobile manipulation, 
integrating a \textbf{multi-view diffusion transformer policy} with a \textbf{reduced whole-body quadratic 
programming (ReM-QP) controller}. 
The framework addresses two long-standing challenges: 
the limited adaptability of classical model-based planners 
and the instability of purely learning-based policies. By leveraging multi-view visual observations and proprioceptive feedback, 
the diffusion transformer policy generates robust high-level end-effector goals. 
The ReM-QP controller then translates these goals into smooth and feasible whole-body motions, 
achieving real-time coordination between the mobile base and the manipulator arm. Extensive evaluations in simulation and real-world environments validate the framework’s effectiveness.  
\lzhang{In simulation, ReM-QP reduces execution time by 28\% and lowers end-effector jerk by 35\%.  
In real-world tasks, \textit{$M^{4}$}Diffuser achieves an average of 82\% success rate with only 6.8\% collisions, 
representing a relative gain of about 28\% higher success and 69\% fewer collisions compared to the OMPL baseline.  
Moreover, our method consistently outperforms strong learning-based baselines, including HoMeR~\cite{sundaresan2025homer}, 
achieving about 10\% higher success and over 5\% fewer collisions.}Generalization experiments further demonstrated robustness for unseen objects and novel scene configurations. Beyond these empirical gains, our framework advances the state of the art in three aspects:  
(1)~it unifies navigation and manipulation through a single whole-body control pipeline, 
avoiding brittle decoupling strategies;  
(2) it removes reliance on fiducial markers, enabling fully marker-free deployment;  
and (3)~it achieves a favorable balance between the adaptability of learning-based policies 
and the precision of optimization-based control, ensuring scalability and deployability. 
In summary, hybridizing diffusion-based visuomotor policies with efficient whole-body optimization 
provides a robust and generalizable solution for mobile manipulation.  
Our framework achieves safe, marker-free operation in unstructured kitchen environments, 
while generalizing to unseen objects and task variations, 
paving the way toward practical deployment in homes, warehouses, 
and healthcare settings. 

Future work will focus on language- and multimodal-guided mobile manipulation.

\vspace{0.1in}

{\small
\bibliographystyle{IEEEtran}
\bibliography{main}
}


\appendix

The supplementary materials include the following:
\begin{itemize}
    \item Implementation details of teleoperation for mobile manipulation in Sec.~\ref{appendix.teleoperation}.
    \item Additional results are shown in Sec.~\ref{appendix.results}.
    
\end{itemize}
\subsection{Implementation Details of Teleoperation}
\label{appendix.teleoperation}

We build our teleoperation pipeline on the \textit{DARKO} robot platform, integrating multiple input devices for intuitive remote control. The overall architecture is organized into three layers: \emph{input}, \emph{control}, and \emph{data flow}.

\textbf{Input Layer:}
At the operator workstation, a 3Dconnexion SpaceMouse provides six degrees of freedom input for end-effector motion commands. These inputs are processed by the ReM-Controller, a whole-body controller running in MuJoCo on a simulated Darko model. The ReM-QP solver computes feasible joint configurations for the mobile base and manipulator, subject to velocity, posture, and collision constraints, ensuring safe and consistent motion generation. In parallel, a standard keyboard is used to control the orientation of a motorized camera mount, handled independently of the whole-body controller.

\textbf{Control Flow:}
The resolved joint commands are transmitted via Wi-Fi from the workstation to the Darko PC, which serves as the central ROS master. The Darko PC dispatches the received commands to the mobile base, the manipulator arm, and the motorized camera mount.

\textbf{Data Flow:}
To manage the bandwidth of the vision sensors, data acquisition is distributed across multiple computation nodes. The NUVO industrial PC collects streams from the front-facing Kinect camera, while a dedicated onboard laptop is directly connected via USB to the rear Kinect and a Realsense camera, thereby avoiding bottlenecks on the NUVO. In addition, the Darko PC aggregates data from multiple LiDAR sensors and forwards them together with the vision streams to the onboard laptop, which serves as the final data center.

\subsection{Additional Results}
\label{appendix.results}
\begin{figure*}[htbp]
    \centering
    \begin{subfigure}[t]{0.4\linewidth}
        \centering
        \includegraphics[width=\linewidth]{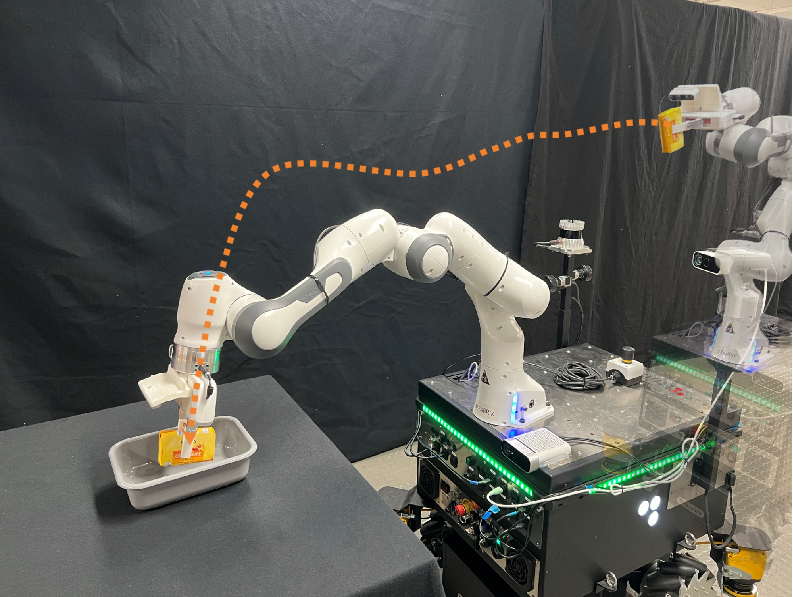}
    \end{subfigure}
    \begin{subfigure}[t]{0.4\linewidth}
        \centering
        \includegraphics[width=\linewidth]{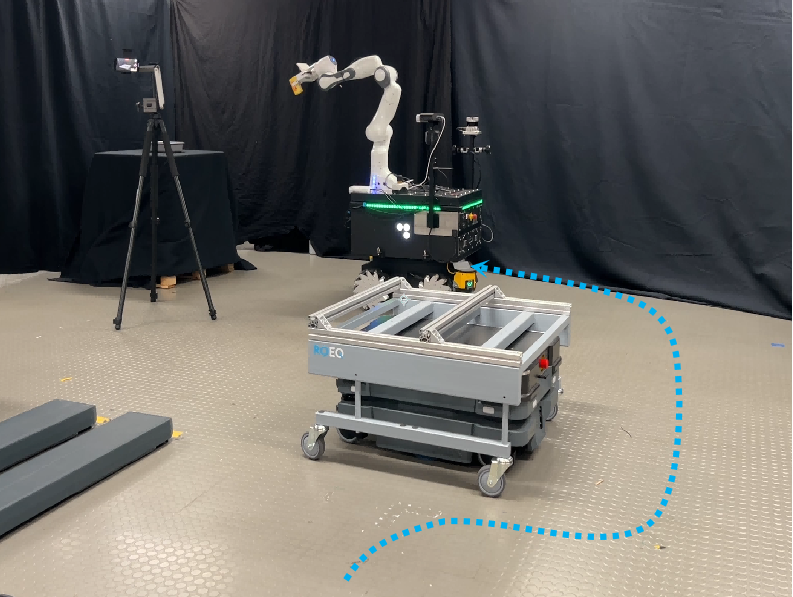}
    \end{subfigure}
    \caption{Trajectory of EE frame (a) and mobile base (b).}
        \label{fig:trajectory_of_ee_and_mobile_base}

\end{figure*}

\begin{figure*}[htbp]
    \centering
        \includegraphics[width=0.8\linewidth]{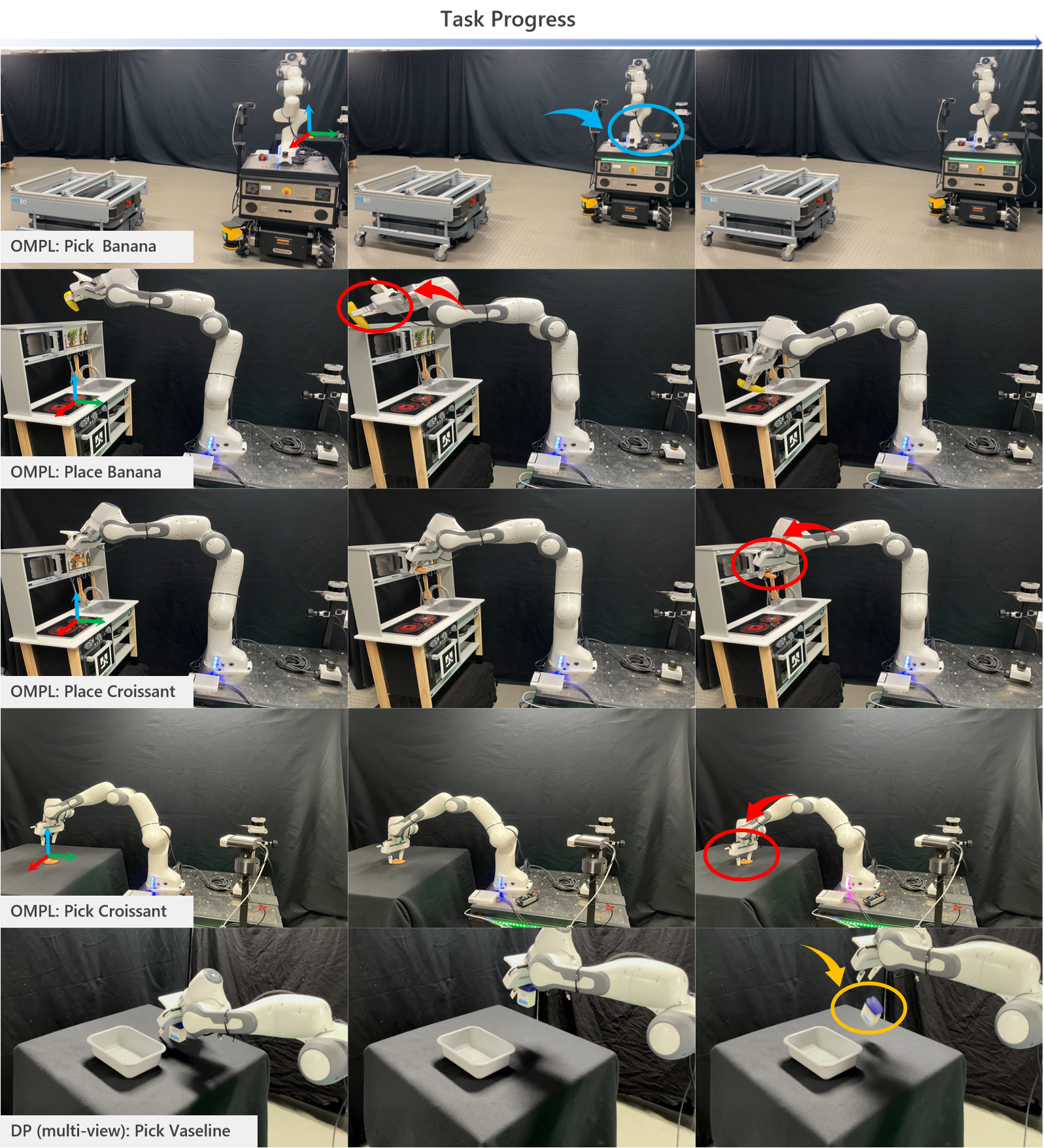}
    \caption{\small Details of the failed experiments.}
    \label{fig:detailed_process_failures}
\end{figure*}

The inference trajectories of EE frame and mobile base are shown in Fig.~\ref{fig:trajectory_of_ee_and_mobile_base}. As well, the addition of the rotatable Kinect joint enhances the overall algorithm’s exploration capability. 
The detailed failure processes are illustrated in Fig.~\ref{fig:detailed_process_failures}.

\end{document}